\documentclass{article}

\usepackage{arxiv}

\usepackage[utf8]{inputenc} 
\usepackage[T1]{fontenc}    
\usepackage{hyperref}       
\usepackage{url}            
\usepackage{booktabs}       
\usepackage{amsfonts}       
\usepackage{nicefrac}       
\usepackage{microtype}      
\usepackage{lipsum}

\usepackage{graphicx}
\usepackage{epsfig}
\usepackage[outdir=./]{epstopdf}
\usepackage[caption=false]{subfig}
\usepackage{multirow}
\usepackage{amsmath}

\usepackage{xcolor}

\usepackage{pifont, newunicodechar}     
\usepackage{array, boldline, makecell}  
\newunicodechar{✓}{\ding{51}}
\newunicodechar{✗}{\ding{55}}

\newcommand{\PreserveBackslash}[1]{\let\temp=\\#1\let\\=\temp}
\newcolumntype{C}[1]{>{\PreserveBackslash\centering}p{#1}}
\newcolumntype{R}[1]{>{\PreserveBackslash\raggedleft}p{#1}}
\newcolumntype{L}[1]{>{\PreserveBackslash\raggedright}p{#1}}

\title{An Ensemble of Simple Convolutional Neural Network Models for MNIST Digit Recognition}

\author{
  Sanghyeon An \quad Minjun Lee \quad Sanglee Park \quad Heerin Yang \quad Jungmin So
  \vspace{0.4cm}\\
  Department of Computer Science and Engineering\\ 
  Sogang University\\
}

\begin{document}
\maketitle

\begin{abstract}
We report that a very high accuracy on the MNIST test set can be achieved by using simple convolutional neural network (CNN) models. We use three different models with 3$\times$3, 5$\times$5, and 7$\times$7 kernel size in the convolution layers. Each model consists of a set of convolution layers followed by a single fully connected layer. Every convolution layer uses batch normalization and ReLU activation,  while pooling is not used. Rotation and translation is used to augment training data, which is a technique frequently used in most image classification tasks. A majority voting using the three models independently trained on the training set can achieve up to 99.87\% accuracy on the test set, which is one of the state-of-the-art results. A two-layer ensemble, a heterogeneous ensemble of three homogeneous ensemble networks, can achieve up to 99.91\% test accuracy. The results can be reproduced by using the code at \href{https://github.com/ansh941/MnistSimpleCNN}{https://github.com/ansh941/MnistSimpleCNN.}
\end{abstract}


\section{Introduction}
\label{sec:1}
\begin{figure}[ht]
	\centering
	\includegraphics[width=0.75\textwidth]{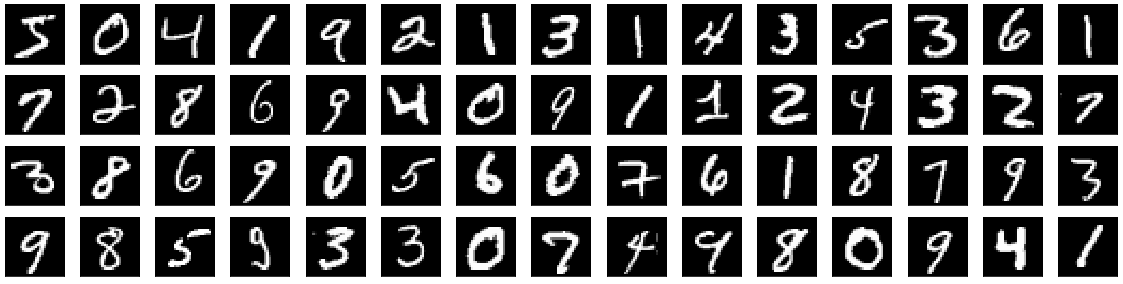}
	\caption{Images from the MNIST training set.}
	\label{fig:mnist}
\end{figure}

MNIST handwritten digit recognition data set (Figure \ref{fig:mnist}, \cite{mnist}) is one of the most basic data sets used to test performance of neural network models and learning techniques. Using 60,000 images as the training set, a 97\%-98\% accuracy could easily be achieved on the test set of 10,000 images, with learning methods such as k-nearest neighbors (KNN), random forests, support vector machines (SVM) and simple neural network models. Convolutional neural networks (CNN) improve this accuracy to over 99\% with less than 100 misclassified images in the test set.

The final 100 images are more difficult to classify correctly. In order to improve accuracy after 99\%, we need more complex models, careful tuning of hyperparameters such as learning rate and batch size, regularization techniques such as batch normalization and dropout, and augmentation of training data. The highest accuracy achieved on the MNIST test set are approximately 99.7\% to 99.84\%, as reported in the papers \cite{byerly20arxiv,kowsari18icisdm,wan13icml,ciresan12cvpr,sabour17nips}.

In this paper, we report a model that can achieve a very high accuracy on the MNIST test set without complex structural aspects or learning techniques. The model uses a set of convolution layers followed by a fully connected layer at the end, which is one of the commonly used model architectures. We use basic data augmentation schemes, translation and rotation. We train three models with similar architectures, and use majority voting between the models to obtain the final prediction. The three models have similar architectures, but have different kernel sizes in the convolution layers. Experiments show that combining models with different kernel sizes achieves better accuracy than combining models with the same kernel size.

\section{Network Design and Training}
\label{sec:2}

Our network models consist of multiple convolution layers and a fully connected layer at the end. In each convolution layer, a 2D convolution is performed, followed by a 2D batch normalization and ReLU activation. Max pooling or average pooling is not used after convolution. Instead, the size of feature map is reduced after each convolution because padding is not used. For example, if we use a 3$\times$3 kernel, the width and height of the image is reduced by two after each convolution layer. Similar approach is taken in other networks \cite{sabour17nips,byerly20arxiv}. The number of channels is increased after each layer in order to account for reduction in feature map size. Once the feature map size becomes small enough, a fully-connected layer connects the feature map to the final output. A 1D batch normalization is used at the fully-connected layer, while dropout is not used.

\begin{figure}[ht]
	\centering
	\subfloat[$M_3$] {\includegraphics[width=0.30\textwidth]{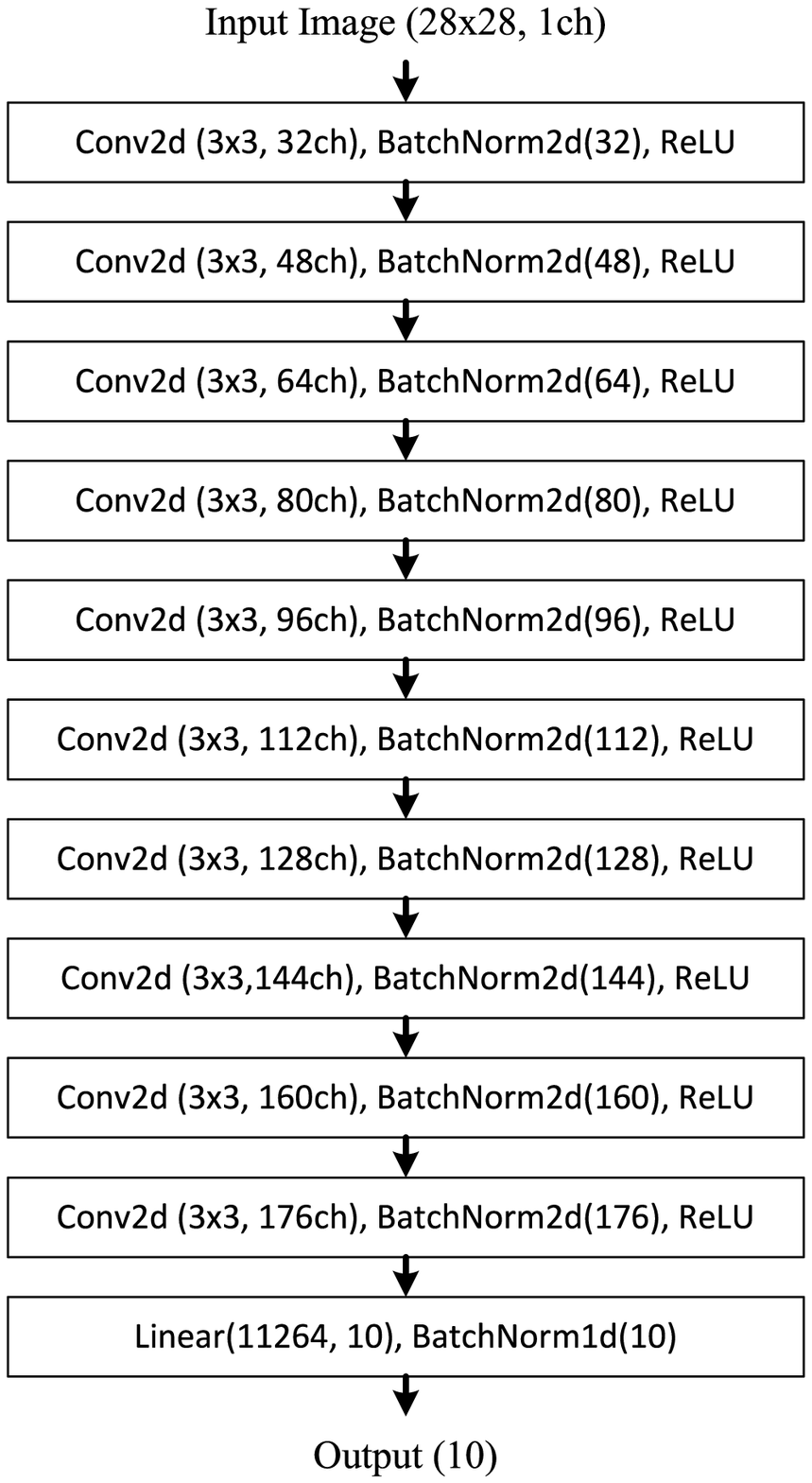}}\quad\quad
	\subfloat[$M_5$] {\includegraphics[width=0.30\textwidth]{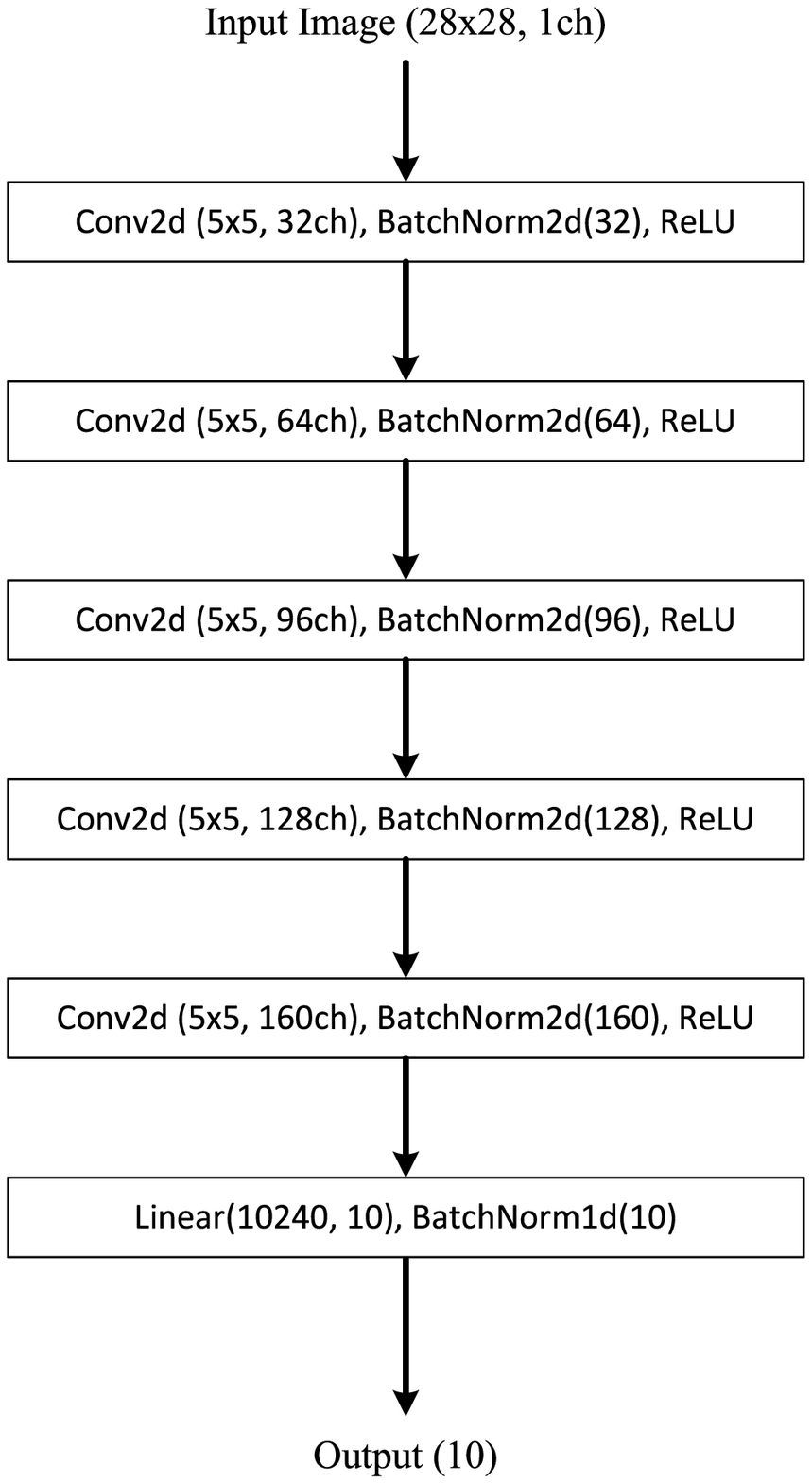}}\quad\quad
	\subfloat[$M_7$] {\includegraphics[width=0.30\textwidth]{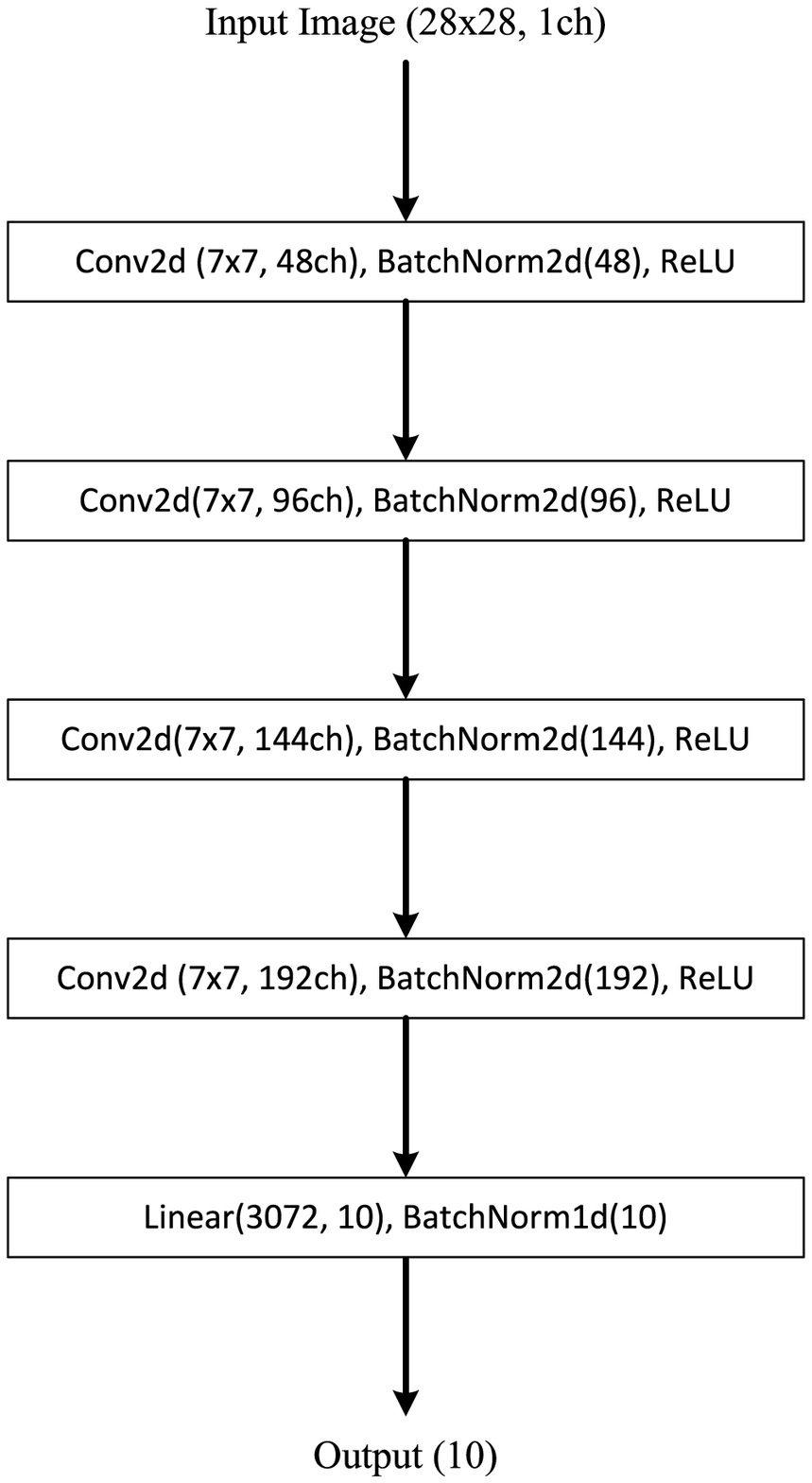}}
	\caption{Network models used for MNIST digit classification.}
	\label{graph:networks}
\end{figure}

We use three different networks and combine the results from these networks. The networks differ only in the kernel sizes of the convolution layers: 3$\times$3, 5$\times$5, and 7$\times$7. Because different kernel size lead to different size reduction in feature maps, the number of layers is different for each network. The first network, $M_3$, uses 10 convolution layers with $16(i+1)$ channels in $i$th convolution layer. The feature map becomes 8$\times$8 with 176 channels after the 10th layer. The second network, $M_5$, uses 5 convolution layers with $32i$ channels in $i$th convolution layer. The feature map becomes 8$\times$8 with 160 channels after the 5th layer. The third network, $M_7$, uses 4 convolution layers with $48i$ channels in $i$th convolution layer. The feature map becomes 4$\times$4 with 192 channels after the 4th layer. The structure of the three networks are shown in Figure \ref{graph:networks}.

\begin{figure}[ht]
	\centering
	\includegraphics[width=0.70\textwidth]{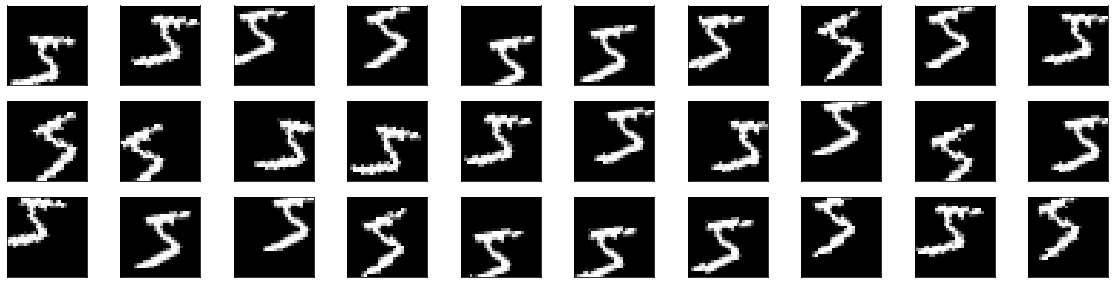}
	\caption{Random translation and random rotation applied to a training image.}
	\label{fig:mnist_transform}
\end{figure}

When training, we apply transformation on data that consist of random translation and random rotation. For random translation, an image is randomly shifted horizontally and vertically, up to 20\% of the image size in each direction. For random rotation, the image is rotated up to 20 degrees in either clockwise or counterclockwise direction. The amount of transformation varies for each image and each epoch, so the network gets to see various versions of an image in the training set (Figure \ref{fig:mnist_transform}). For training and evaluation, the input vectors which are typically integers in [0, 255] are converted to floating point values in [-1.0, 1.0].

The network parameters are initialized using default initialization methods in PyTorch \cite{paszke19nips}. For parameter optimization, we use the Adam optimizer with cross-entropy loss function. Learning rate starts at 0.001, and exponentially decays with decaying factor $\gamma$=0.98. The batch size is 120, and so 500 parameter updates occur in an epoch. We use exponential moving average of weights for evaluation, which may lead to better generalization \cite{izmailov18uai}. The exponential decay used for computing the moving average is 0.999.

\section{Experiments}

\subsection{Results for Individual Networks and Ensembles}
\begin{figure}[ht]
	\centering
	\subfloat[$M_3$] {\includegraphics[width=0.33\textwidth]{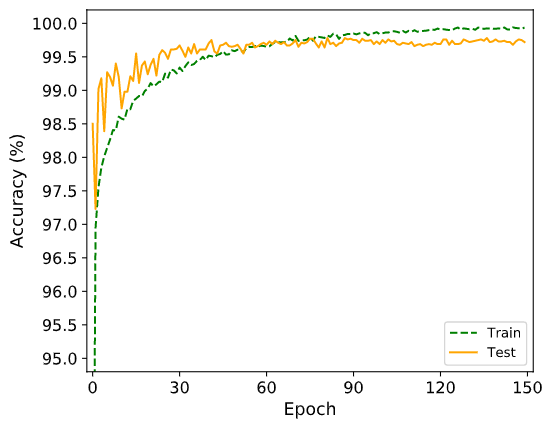}}
	\subfloat[$M_5$] {\includegraphics[width=0.33\textwidth]{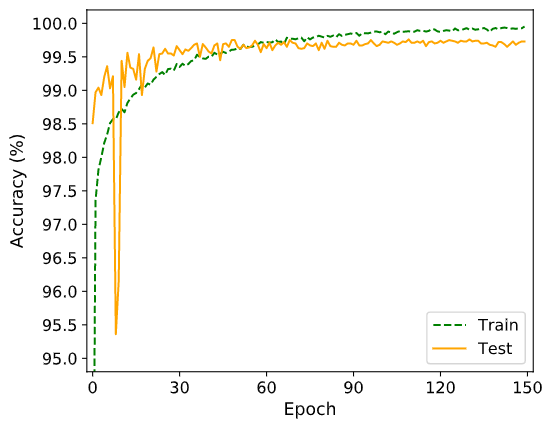}}
	\subfloat[$M_7$] {\includegraphics[width=0.33\textwidth]{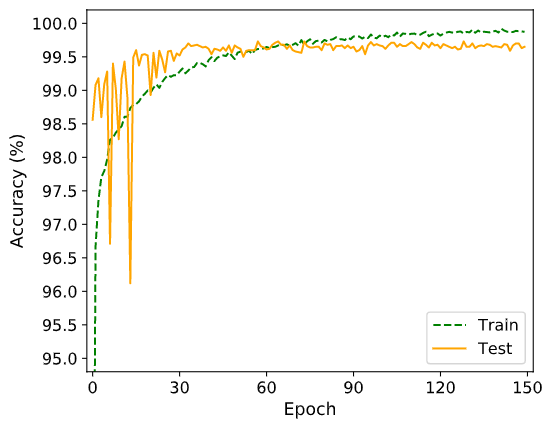}}
	\caption{Train accuracy and test accuracy of $M_3$, $M_5$, and $M_7$ during training.}
	\label{graph:training}
\end{figure}

For each type of network, we have trained 30 networks with different initial parameters. Each network was trained for 150 epochs, since the test accuracy hardly improved after that point. Figure \ref{graph:training} shows the change in the training accuracy and the test accuracy while training. In terms of test accuracy, networks with larger kernels show some instability at early epochs, but the patterns of all networks become similar after 50 epochs. Table \ref{tab:training} shows the minimum, average, maximum accuracy of 30 networks between 50 and 150 epochs, in the 95\% confidence range. The accuracy of $M_3$ is slightly higher followed by $M_5$ and $M_7$, but the difference is not too significant (less than 0.02\%). Between 50 and 150 epochs of 30 networks, the highest test accuracy observed from $M_3$, $M_5$, $M_7$ was 99.82, 99.80, and 99.79 respectively.

\begin{table}[ht]
 \caption{Test accuracy of networks measured between 50 epoch and 150 epoch in training.}
  \centering
  \begin{tabular}{C{0.6in}cccC{0.6in}}
    \toprule
    \multirow{2}{*}{model} & \multicolumn{4}{c}{test accuracy}          \\
    \cmidrule(r){2-5}
          & min                   & avg                     & max                   & best  \\
    \toprule
    $M_3$ & 99.5930 $\pm$ 0.0136  & 99.6949 $\pm$ 0.0058    & 99.7667 $\pm$ 0.0084  & 99.82 \\ 
    \midrule
    $M_5$ & 99.5863 $\pm$ 0.0115  & 99.6835 $\pm$ 0.0074    & 99.7583 $\pm$ 0.0081  & 99.80 \\
    \midrule
    $M_7$ & 99.5470 $\pm$ 0.0288  & 99.6711 $\pm$ 0.0089    & 99.7450 $\pm$ 0.0093  & 99.79 \\
    \bottomrule
  \end{tabular}
  \label{tab:training}
\end{table}

It is known that using ensemble of networks can improve generalization and achieve higher test accuracy \cite{breiman96ml,freund00stat,friedman01stat,ke17nips}. To test the performance of ensemble networks on the MNIST data set, we trained 30 networks each of $M_3$, $M_5$, and $M_7$, and tested four different ensemble strategies. In the first three strategies, we randomly select three networks from the same type of networks ($M_3$, $M_5$, or $M_7$). In the fourth strategy, we select one network from each type. The final result is obtained by using majority voting. That is, if two networks agree that an image belongs to a particular class, that class is selected. If the three networks vote on different class, one class is randomly selected among the three. For each strategy, we tested 1000 ensemble networks and plotted the histogram for the test accuracy. 

\begin{figure}[ht]
	\centering
	\subfloat[$M_3$] {\includegraphics[width=0.33\textwidth]{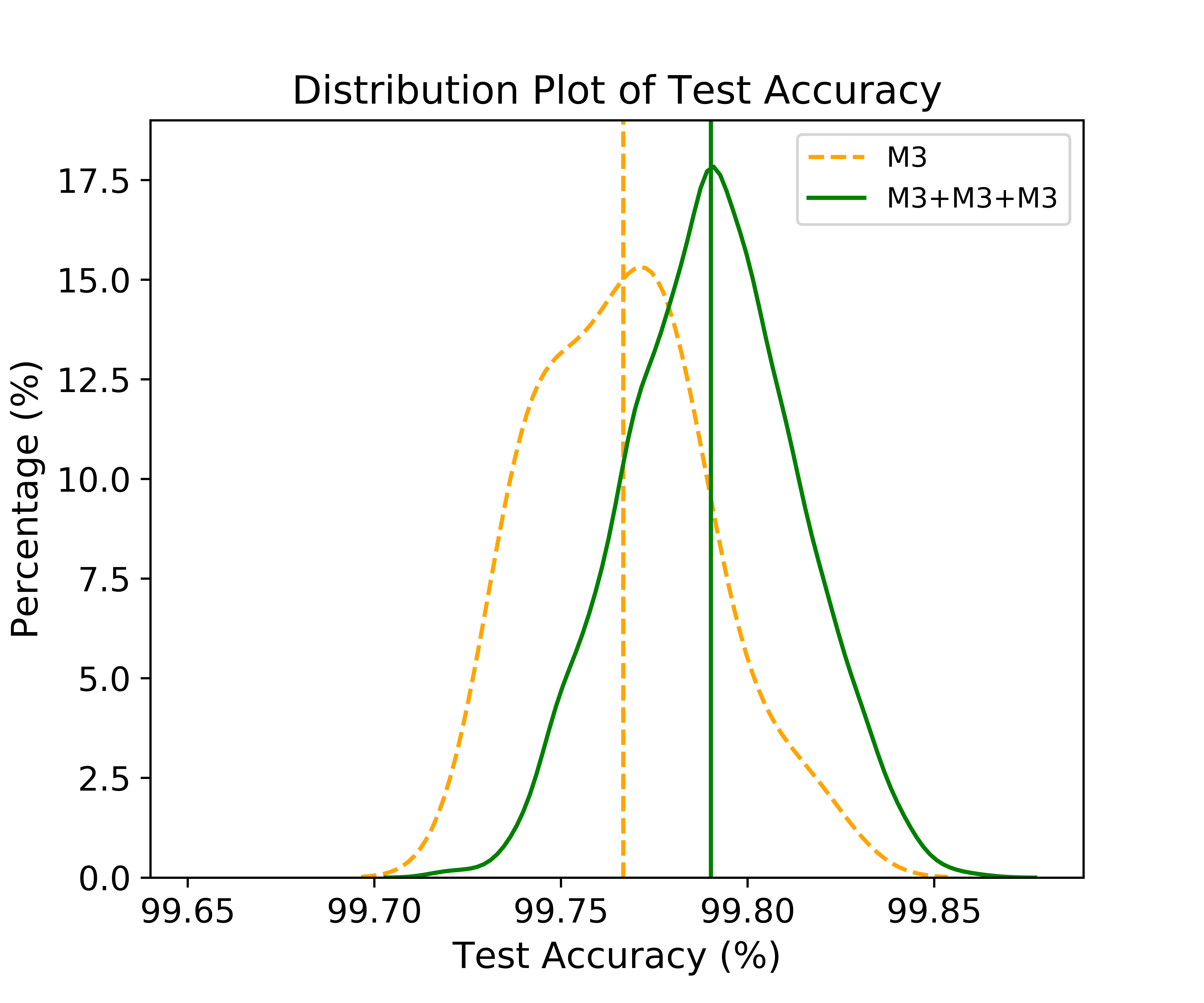}}
	\subfloat[$M_5$] {\includegraphics[width=0.33\textwidth]{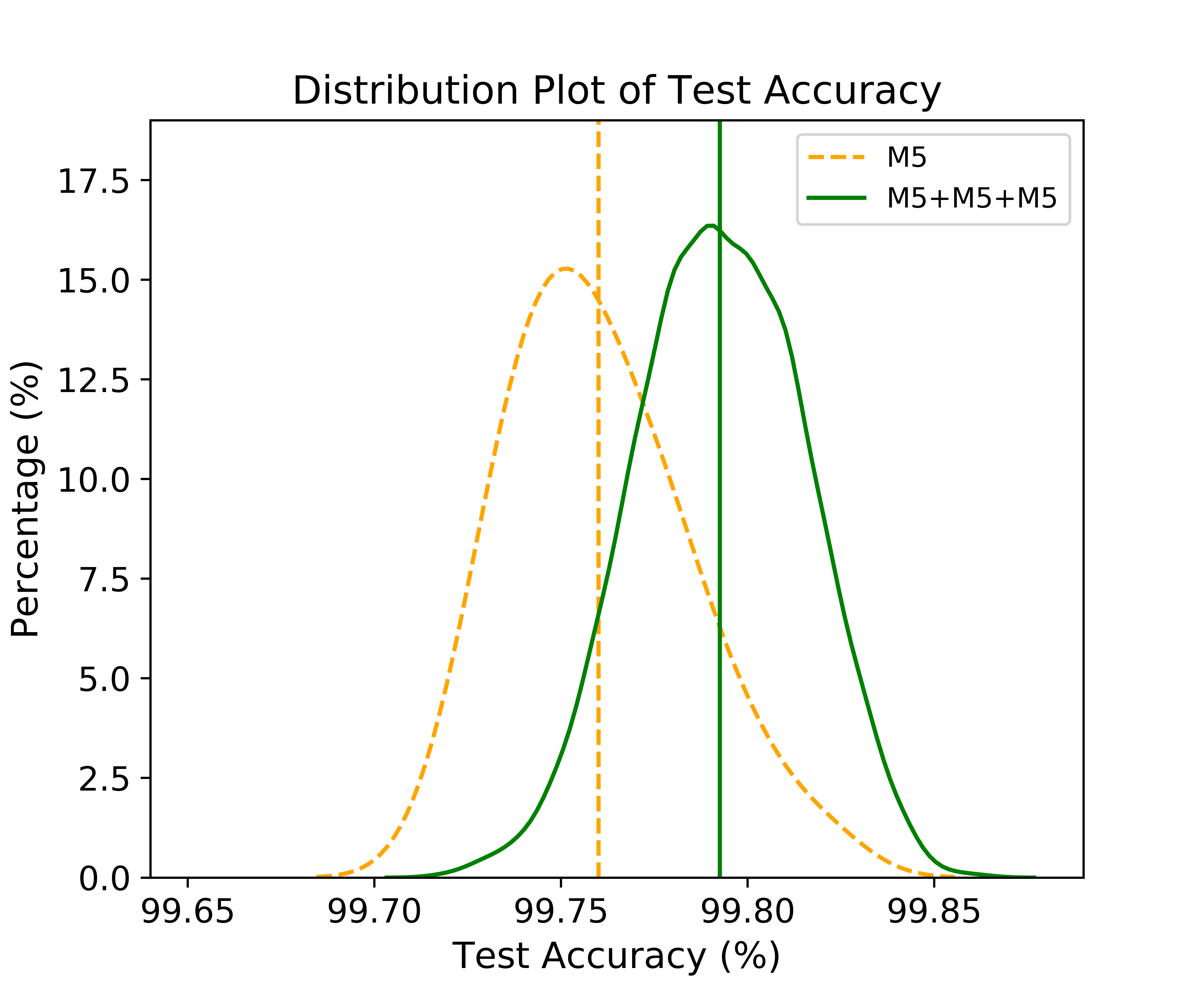}}
	\subfloat[$M_7$] {\includegraphics[width=0.33\textwidth]{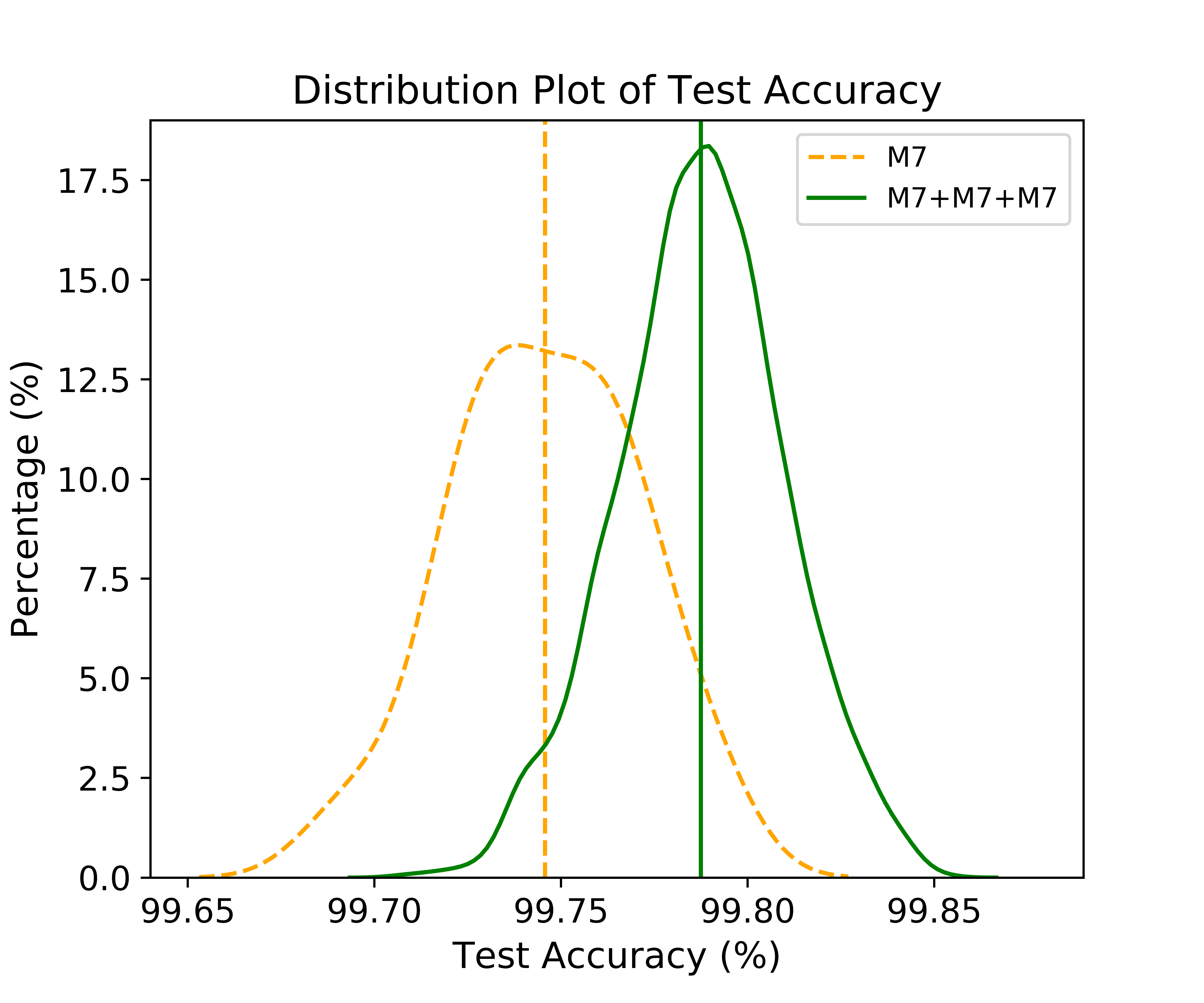}}
	\caption{Distribution of test accuracy for individual networks and homogeneous ensemble networks.}
	\label{graph:comparison}
\end{figure}

\begin{figure}[ht]
    \centering
    \includegraphics[width=0.66\textwidth]{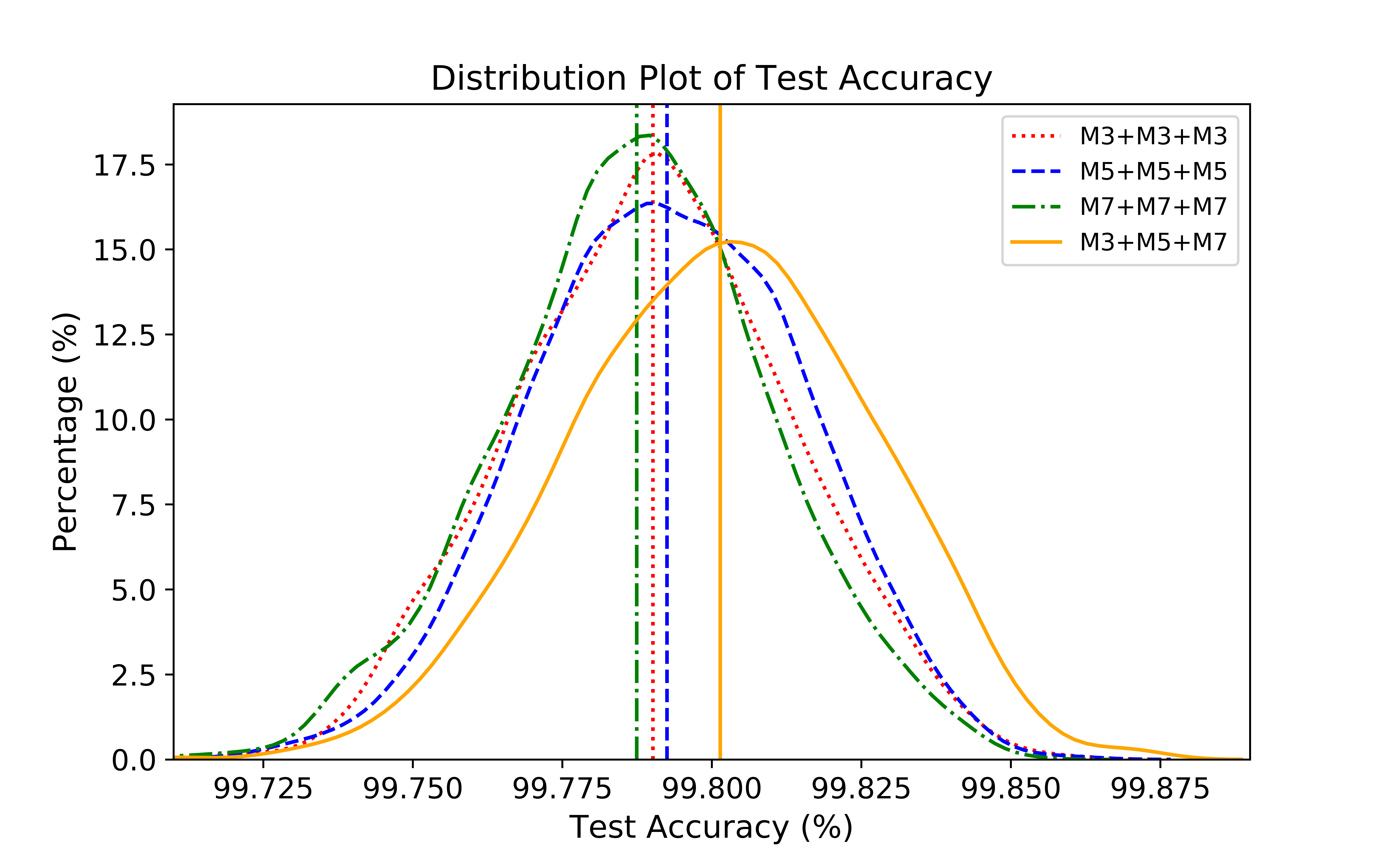}
    \caption{Distribution of test accuracy for homogeneous and heterogeneous ensemble networks.}
    \label{graph:combined}
\end{figure}

Figure \ref{graph:comparison} shows the benefit of using ensemble of homogeneous networks. For $M_3$, $M_5$, and $M_7$, higher test accuracy could be achieved by combining results from three networks. (The line moves to the right.) Figure \ref{graph:combined} shows the test accuracy of the four ensemble methods discussed above, and Table \ref{tab:ensemble} shows the 95\% confidence range of test accuracy for the four methods. It can be observed that while the average test accuracy of homogeneous ensemble methods are similar, the ensemble method where one network is selected from each type of networks achieves higher accuracy.

\begin{table}[ht]
 \caption{95\% confidence range of test accuracy for homogeneous and heterogeneous ensemble networks.}
  \centering
  \begin{tabular}{C{1.2in}C{1.4in}C{0.8in}}
    \toprule
    \multirow{2}{*}{configuration} & \multicolumn{2}{c}{test accuracy}          \\
    \cmidrule(r){2-3}
                    & 95\% confidence range & best accuracy \\
    \toprule
    $M_3+M_3+M_3$   & 99.7901 $\pm$ 0.0014 & 99.86\\
    \midrule
    $M_5+M_5+M_5$   & 99.7925 $\pm$ 0.0014 & 99.86\\
    \midrule
    $M_7+M_7+M_7$   & 99.7874 $\pm$ 0.0014 & 99.85\\
    \midrule
    $M_3+M_5+M_7$   & \textbf{99.8014} $\pm$ \textbf{0.0015} & \textbf{99.87}\\
    \bottomrule
  \end{tabular}
  \label{tab:ensemble}
\end{table}

\begin{figure}[ht]
	\centering
	\includegraphics[width=0.66\textwidth]{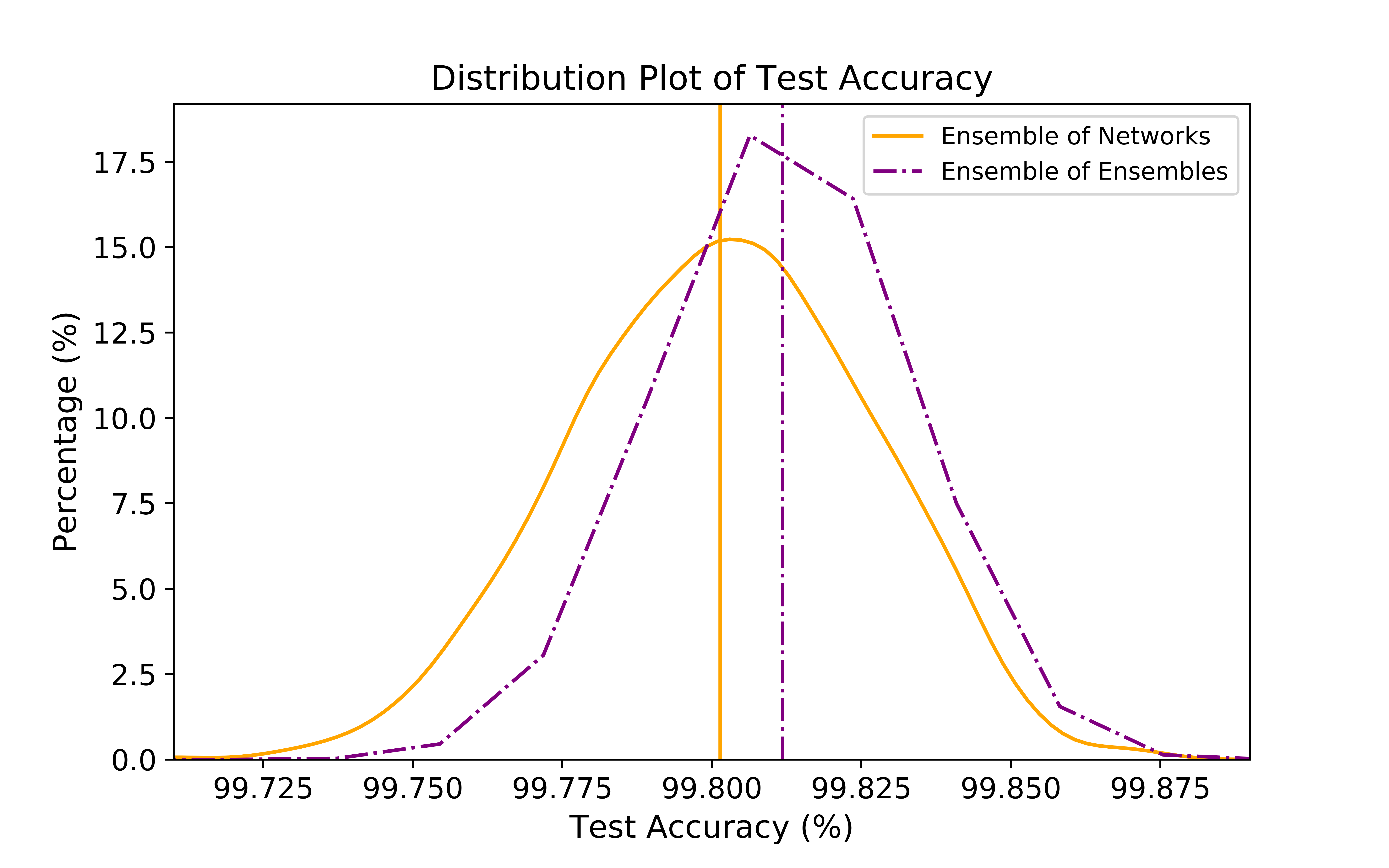}
	\caption{Distribution of test accuracy for ensemble of individual networks and ensemble of ensemble networks.}
	\label{graph:ensemble2}
\end{figure}

From Figure \ref{graph:comparison} we can see that using an ensemble of three homogeneous networks could improve the test accuracy. Also, it is shown in Figure \ref{graph:combined} that combining results from heterogeneous networks could help boost the accuracy as well. We have tested a two-level ensemble method, where we first combine results from three homogeneous networks, and then combine results from three homogeneous ensemble networks. For this study, we trained 3 groups of 10 networks for each type of network, $M_3$, $M_5$, and $M_7$. For each network, we trained for 150 epochs and saved the best model in terms of test accuracy. Then, we randomly chose 3 networks from $M_3$ and combined their results using majority voting. Similarly, we combined results of three networks for $M_5$ and $M_7$. After that, we used majority voting for the three ensemble networks. Figure \ref{graph:ensemble2} shows the distribution of test accuracy for 1000 ensemble of individual networks ($M_3$+$M_5$+$M_7$) and 1000 ensemble of ensemble networks (($M_3$+$M_3$+$M_3$)+($M_5$+$M_5$+$M_5$)+($M_7$+$M_7$+$M_7$)). The graph shows that using ensemble of ensemble networks improves the test accuracy in average. Table \ref{tab:ensemble2} shows the 95\% confidence range and the best accuracy observed for ensemble of individual and ensemble of ensemble networks. In addition to random selection, we also show the best case in order to see what is the best accuracy we can achieve. For the best case, we picked 10 homogeneous ensemble networks from $M_3$, $M_5$, and $M_7$ that shows the best test accuracy. Then, we chose one network from each type and combined their results. The best accuracy achieved was 99.91\%.

\begin{table}[ht]
 \caption{95\% confidence range and best-case test accuracy for ensemble of individual networks and ensemble of ensemble networks.}
  \centering
  \begin{tabular}{C{1.8in}C{1.4in}C{0.8in}}
    \toprule
    \multirow{2}{*}{configuration} & \multicolumn{2}{c}{test accuracy}          \\
    \cmidrule(r){2-3}
                              & 95\% confidence range & best accuracy \\
    \toprule
    ensemble of individual networks & 99.8014 $\pm$ 0.0015    & 99.87 \\
    \midrule
    ensemble of ensembles (random)  & 99.8118 $\pm$ 0.0002    & 99.89 \\
    \midrule
    ensemble of ensembles (best)    & 99.8646 $\pm$ 0.0008    & 99.91 \\
    \bottomrule
  \end{tabular}
  \label{tab:ensemble2}
\end{table}

\subsection{Impact of network architecture}
\begin{figure}[ht]
	\centering
	\subfloat[$C_1$] {\includegraphics[width=0.30\textwidth]{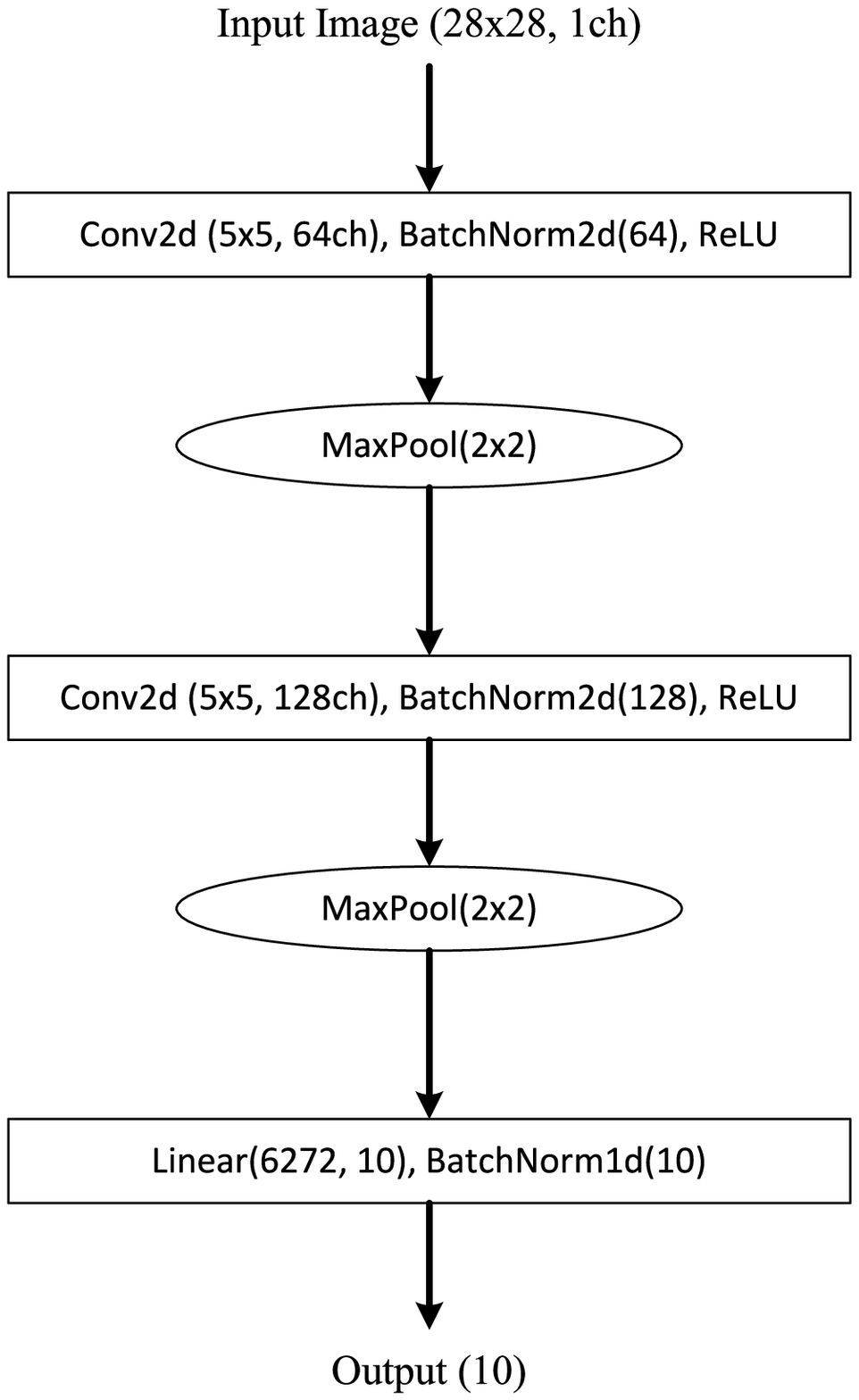}}\quad\quad
	\subfloat[$C_2$] {\includegraphics[width=0.30\textwidth]{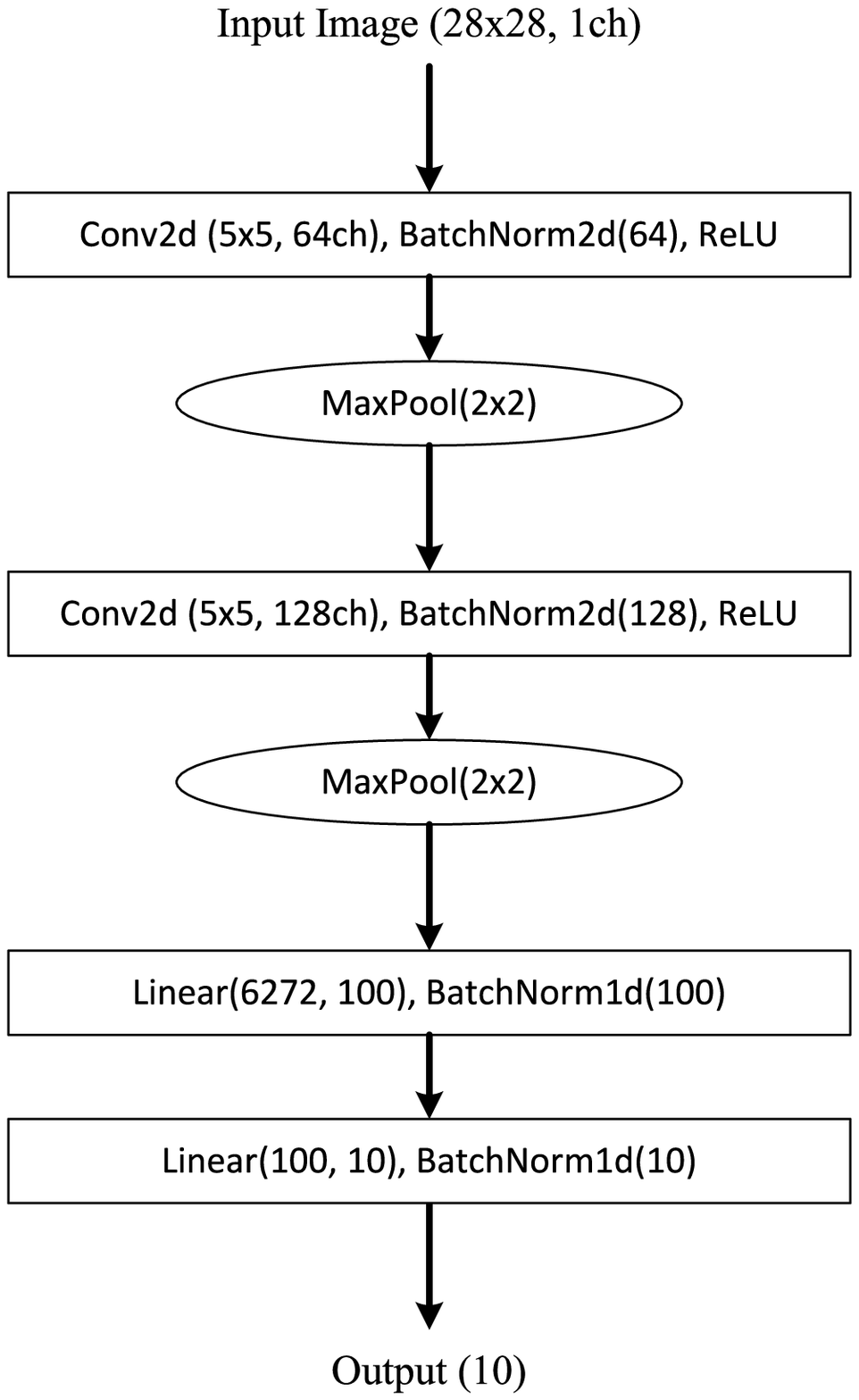}}\quad\quad
	\subfloat[$C_3$] {\includegraphics[width=0.30\textwidth]{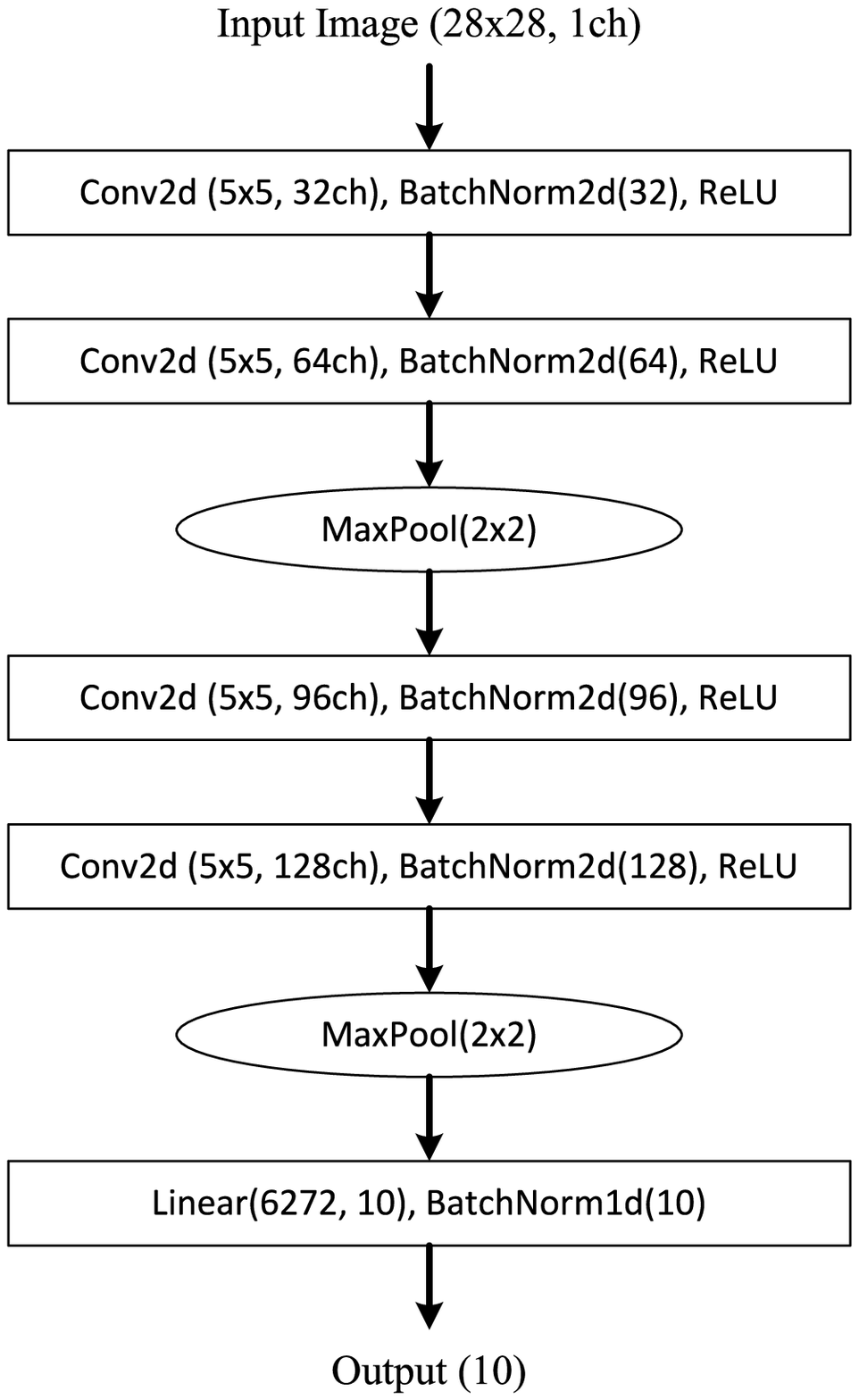}}
	\caption{Commonly used CNN structures with max pooling.}
	\label{fig:comparenetworks}
\end{figure}

When building a CNN, a common practice is to use pooling, such as max pooling or average pooling \cite{ranzato07cvpr}. Pooling is used to obtain translation invariance and also reduce dimension of the feature maps. A commonly used CNN model consists of a set of convolution layers where each convolution layer is followed by a pooling layer, and one or multiple fully connected layers at the end. Some networks have two convolution layers before the pooling layer. Figure \ref{fig:comparenetworks} show some of the commonly used CNN structures, and we name the three networks $C_1$, $C_2$, and $C_3$.

\begin{figure}[ht]
	\centering
	\subfloat[$C_1$] {\includegraphics[width=0.25\textwidth]{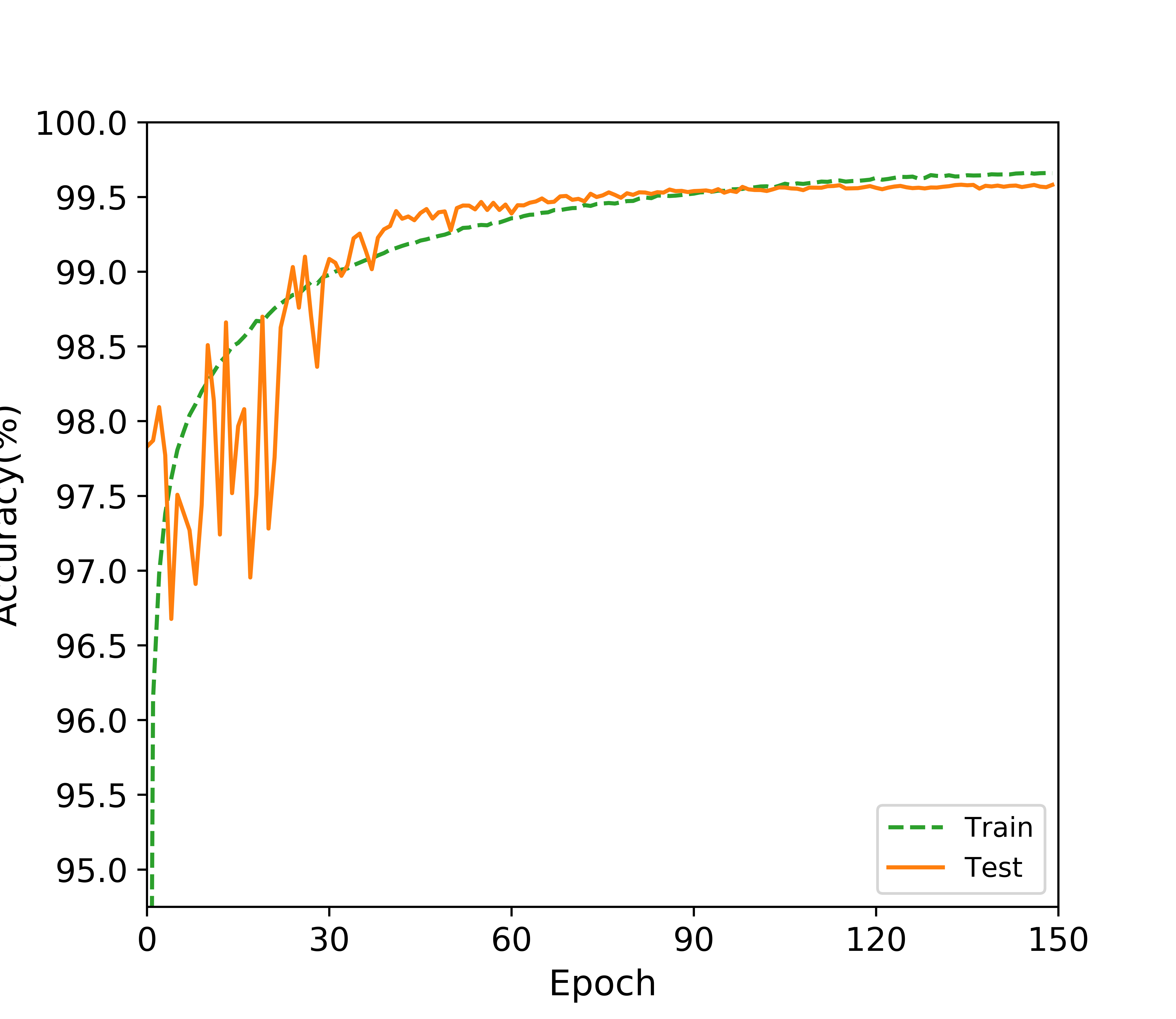}}
	\subfloat[$C_2$] {\includegraphics[width=0.25\textwidth]{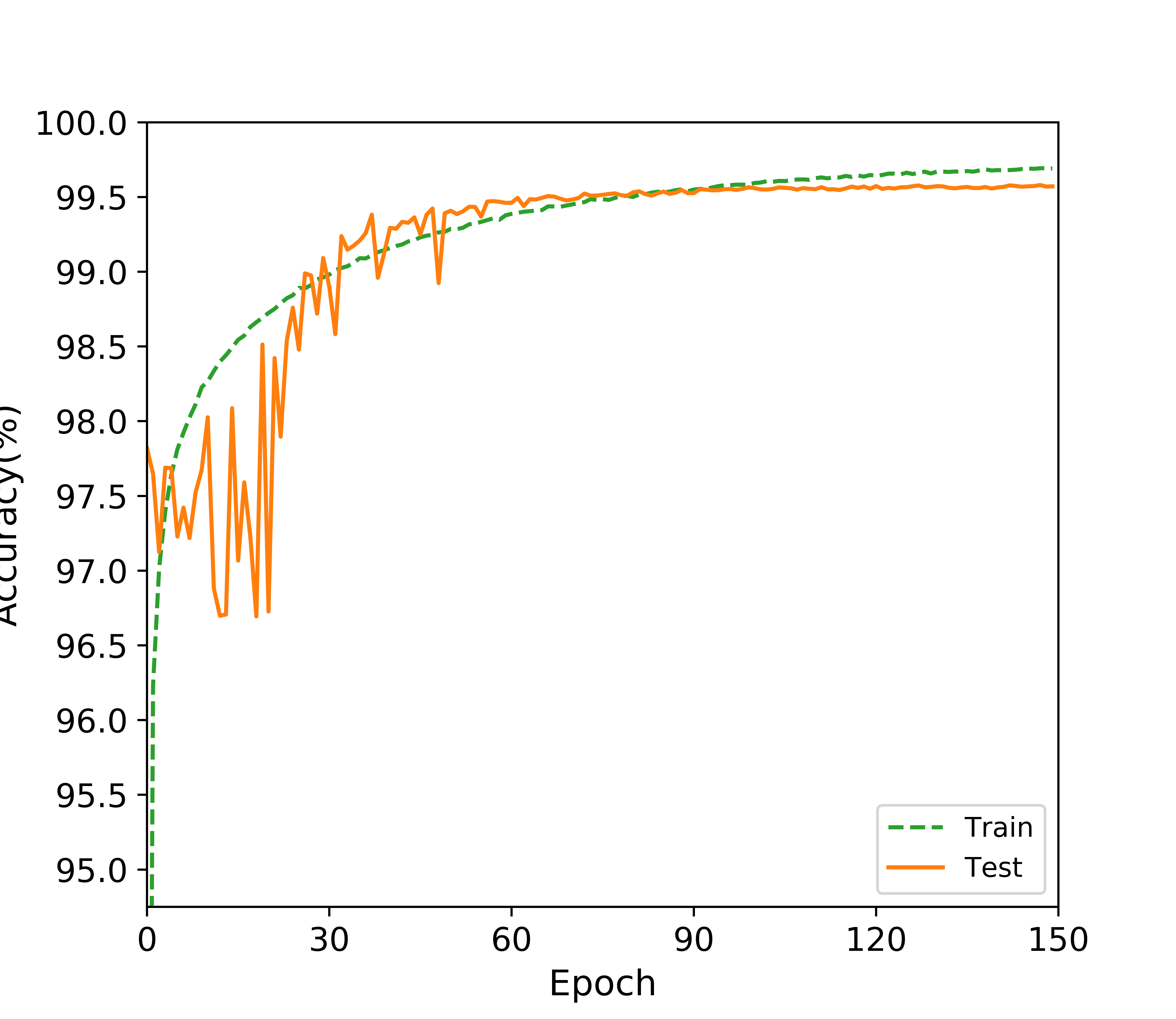}}
	\subfloat[$C_3$] {\includegraphics[width=0.25\textwidth]{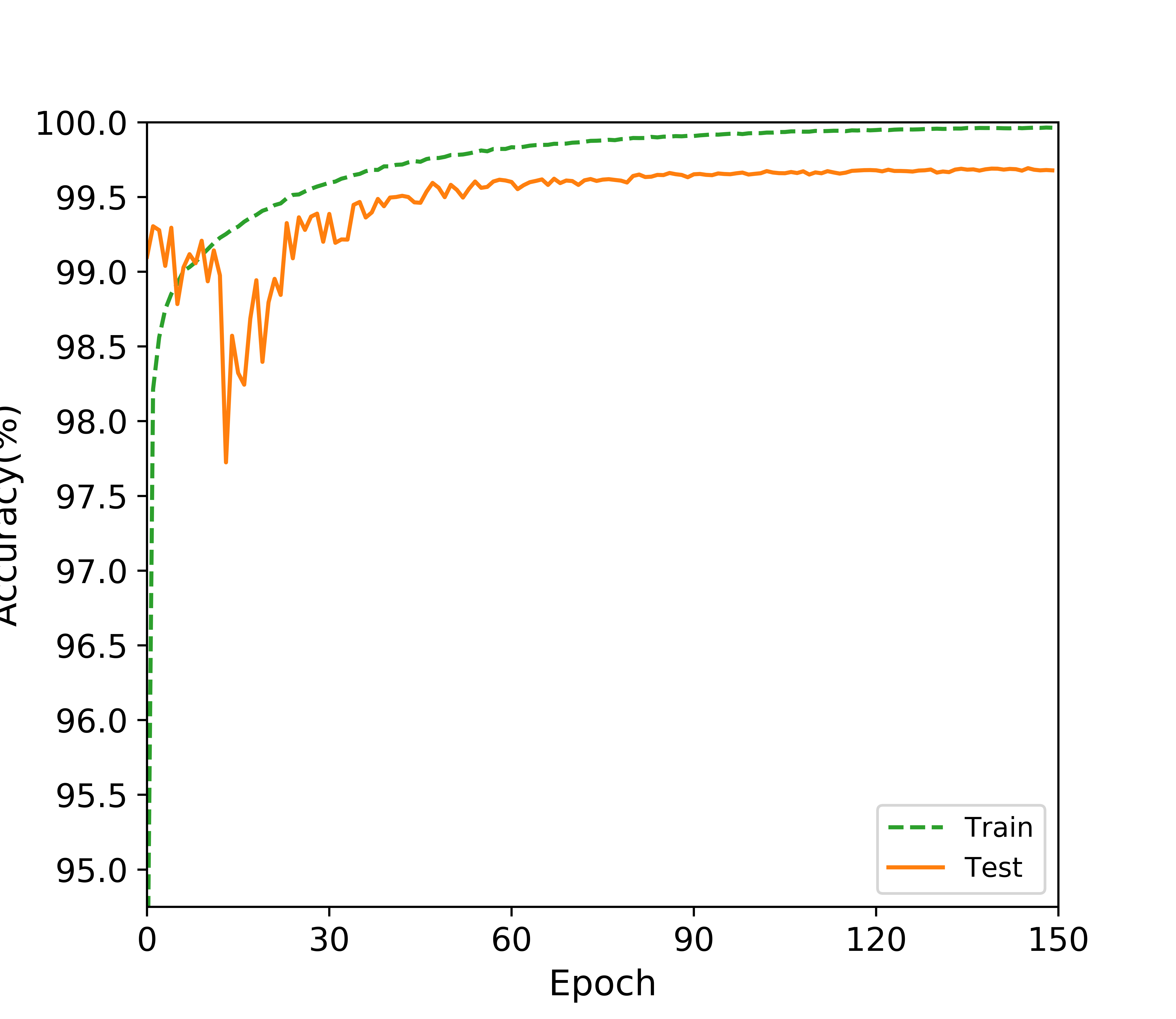}}
	\subfloat[$M_5$] {\includegraphics[width=0.25\textwidth]{3_1/accuracy_m5.png}}
	\caption{Train accuracy and test accuracy of $C_1$, $C_2$, $C_3$, and $M_5$ during training.}
	\label{graph:maxpoolcomparison2}
\end{figure}

Figure \ref{graph:maxpoolcomparison2} shows the change in training and test accuracy during training. It can be observed that for networks using max pooling, the test accuracy goes through oscillations in the early stage of training. On the other hand, the test accuracy of $M_5$ increases in a more stable manner. Table \ref{table:maxpoolcomparison2_2} shows the test accuracy of 30 networks between 50 epoch and 150 epoch of training. The average test accuracy of $C_3$ and $M_5$ is better than that of $C_1$ and $C_2$, which means using more convolution layers could result in better feature learning. Having more fully connected layers at the end did not help, as can be seen from the accuracy of $C_1$ and $C_2$. Between $C_3$ and $M_5$, $M_5$ achieves higher accuracy in general, and also can reach higher accuracy in the best case.

\begin{table}[ht]
 \caption{Test accuracy of networks measured between 50 epoch and 150 epoch in training.}
  \centering
  \begin{tabular}{C{0.6in}cccC{0.6in}}
    \toprule
    \multirow{2}{*}{model} & \multicolumn{4}{c}{test accuracy}          \\
    \cmidrule(r){2-5}
          & min                   & avg                     & max                   & best  \\
    \toprule
    $C_1$ & 99.3052 $\pm$ 0.0865  & 99.5293 $\pm$ 0.0105    & 99.6419 $\pm$ 0.0059  & 99.70 \\ 
    \midrule
    $C_2$ & 99.3594 $\pm$ 0.0442  & 99.5316 $\pm$ 0.0090    & 99.6337 $\pm$ 0.0051  & 99.68 \\
    \midrule
    $C_3$ & 99.4720 $\pm$ 0.04268  & 99.6448 $\pm$ 0.0078    & 99.7372 $\pm$ 0.0033  & 99.78 \\
    \midrule
    $M_5$ & \textbf{99.5863 $\pm$ 0.0115} & \textbf{99.6835 $\pm$ 0.0074} & \textbf{99.7583 $\pm$ 0.0081} & \textbf{99.80} \\
    \bottomrule
  \end{tabular}
  \label{table:maxpoolcomparison2_2}
\end{table}

\begin{figure}[ht]
	\centering
	\includegraphics[width=0.62\textwidth]{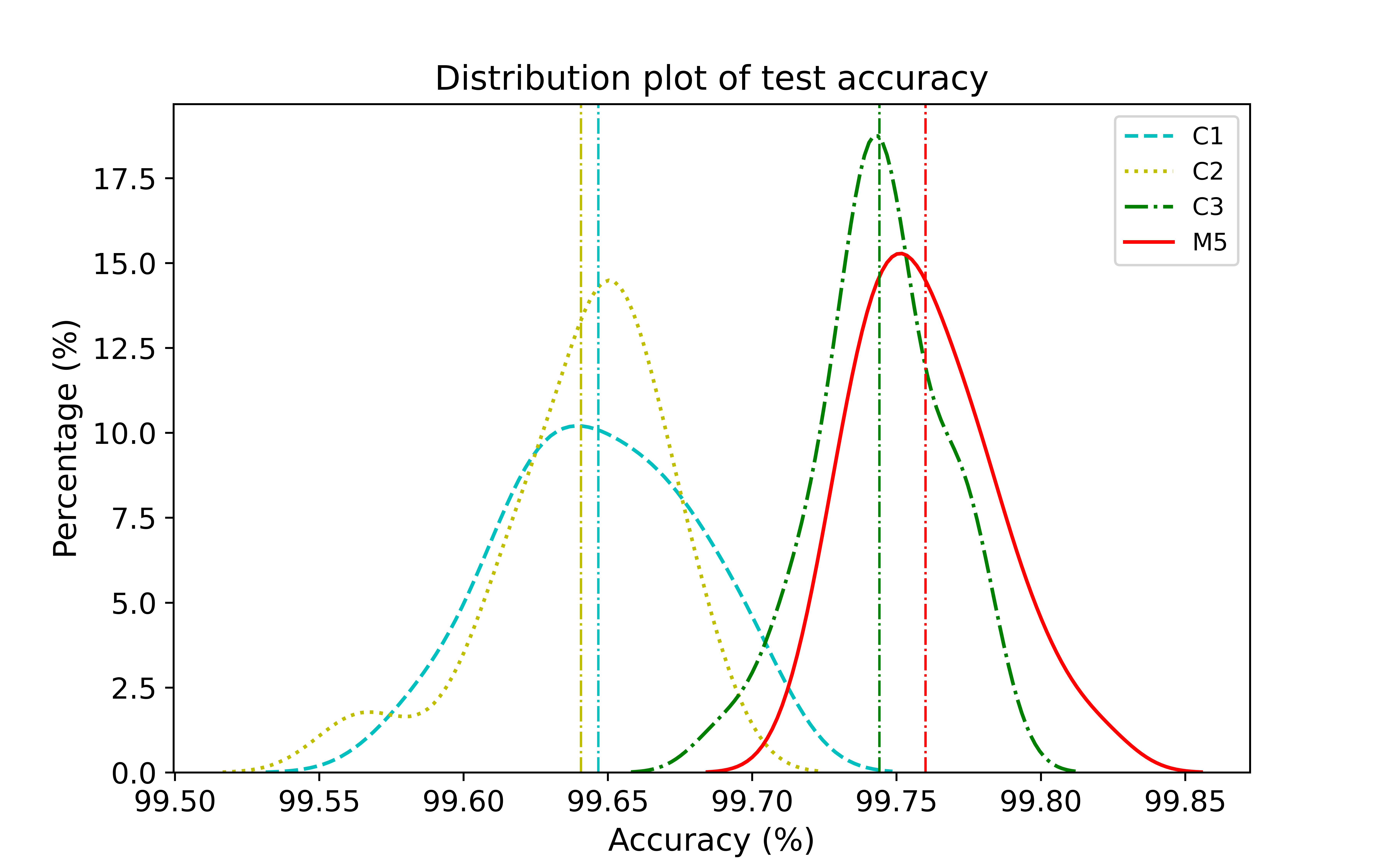}
	\caption{Distribution of test accuracy for networks with different architectures.}
	\label{graph:maxpoolcomparison1}
\end{figure}

Figure \ref{graph:maxpoolcomparison1} shows the distribution plot of 30 networks for $C_1$, $C_2$, $C_3$, and $M_5$. For this graph, each network is trained for 150 epochs, and the network with the highest test accuracy is saved. It can be shown that $M_5$ achieves better test accuracy than other networks in general. Table \ref{table:maxpoolcomparison2_1} shows the 95\% confidence range of test accuracy for each network models.

\begin{table}[ht]
    \centering
    \caption{95\% confidence range of test accuracy for networks with different architectures.}
    \begin{tabular}{C{0.8in}C{1.6in}}
    \toprule
    model             & test accuracy                              \\
    \toprule
    $C_1$             & 99.6466 $\pm$ 0.0121                       \\
    \midrule
    $C_2$             & 99.6406 $\pm$ 0.0108                       \\
    \midrule
    $C_3$             & 99.7440 $\pm$ 0.0080                       \\
    \midrule
    $M_5$             & \textbf{99.7600} $\pm$ \textbf{0.0089}     \\
    \bottomrule
    \end{tabular}
    \label{table:maxpoolcomparison2_1}
\end{table}

\subsection{Impact of data augmentation}

\begin{figure}[ht]
    \centering
    \includegraphics[width = 0.61\textwidth]{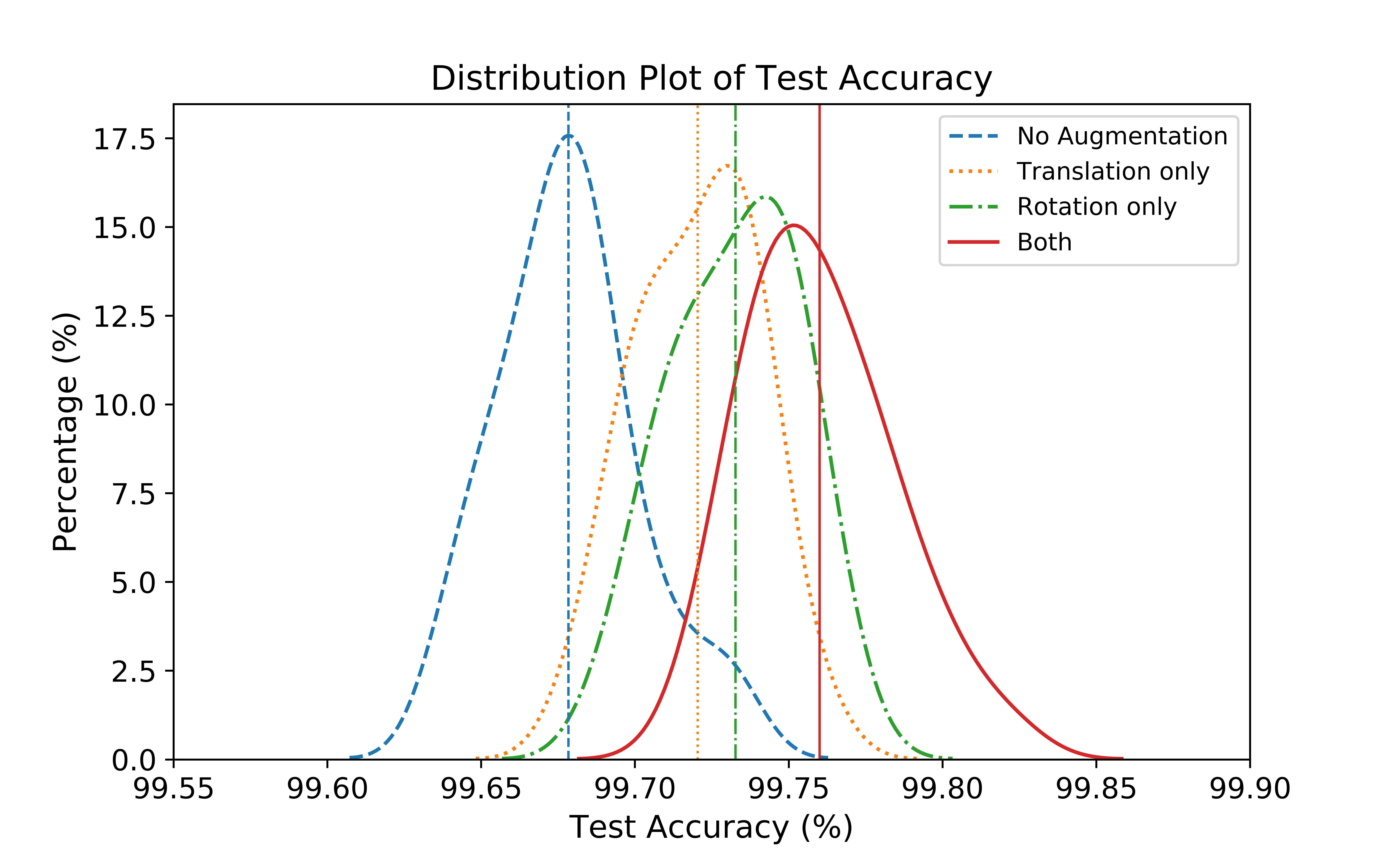}
    \caption{Distribution of test accuracy for networks trained with different augmentation schemes.}
    \label{graph:Aug_frequency}
\end{figure}

Data augmentation is a technique to increase the diversity of training data without actually collecting data and labeling them. It is an essential technique for supervised learning in which a large data set is required for the network model to achieve high performance \cite{cubuk19cvpr,cubuk19arxiv,zhong17arxiv,kim19arxiv,bachman19arxiv}. When training the proposed network, we used two schemes for data generation: random rotation and random translation. There are many other schemes such as cropping, flipping, and resizing, and the best augmentation schemes depend on the data. In this section, we study whether data augmentation actually helps improving the network performance. We compared performance of four $M_5$ networks with different combinations of augmentation schemes applied. Figure \ref{graph:Aug_frequency} shows the distribution plot of 30 networks for four different augmentation strategies. It can be observed that data augmentation is helpful in general. For the MNIST data set, applying random rotation has slightly higher contribution than random translation, but both schemes are needed to achieve the best accuracy. Table \ref{Table:Aug} shows the 95\% confidence range of test accuracy for the four augmentation strategies.


\begin{table}[ht]
    \centering
    \caption{95\% confidence range of test accuracy for networks trained with different augmentation schemes.}
    \label{Table:Aug}
    \begin{tabular}{C{1.0in}C{1.0in}C{1.6in}} 
        \toprule
        \multicolumn{2}{c}{augmentation scheme} & \multirow{2}{*}{test accuracy} \\
        \cmidrule(r){1-2}
        translation & rotation &    \\
        \toprule
        ✗ & ✗ & 99.6783 $\pm$ 0.0086 \\ 
        \midrule
        ✓ & ✗ & 99.7203 $\pm$ 0.0074 \\ 
        \midrule
        ✗ & ✓ & 99.7327 $\pm$ 0.0077 \\ 
        \midrule
        ✓ & ✓ & \textbf{99.7600} $\pm$ \textbf{0.0089} \\
        \bottomrule
    \end{tabular}
\end{table}


\subsection{Impact of batch normalization}

\begin{figure}[ht]
	\centering
	\includegraphics[width=0.61\textwidth]{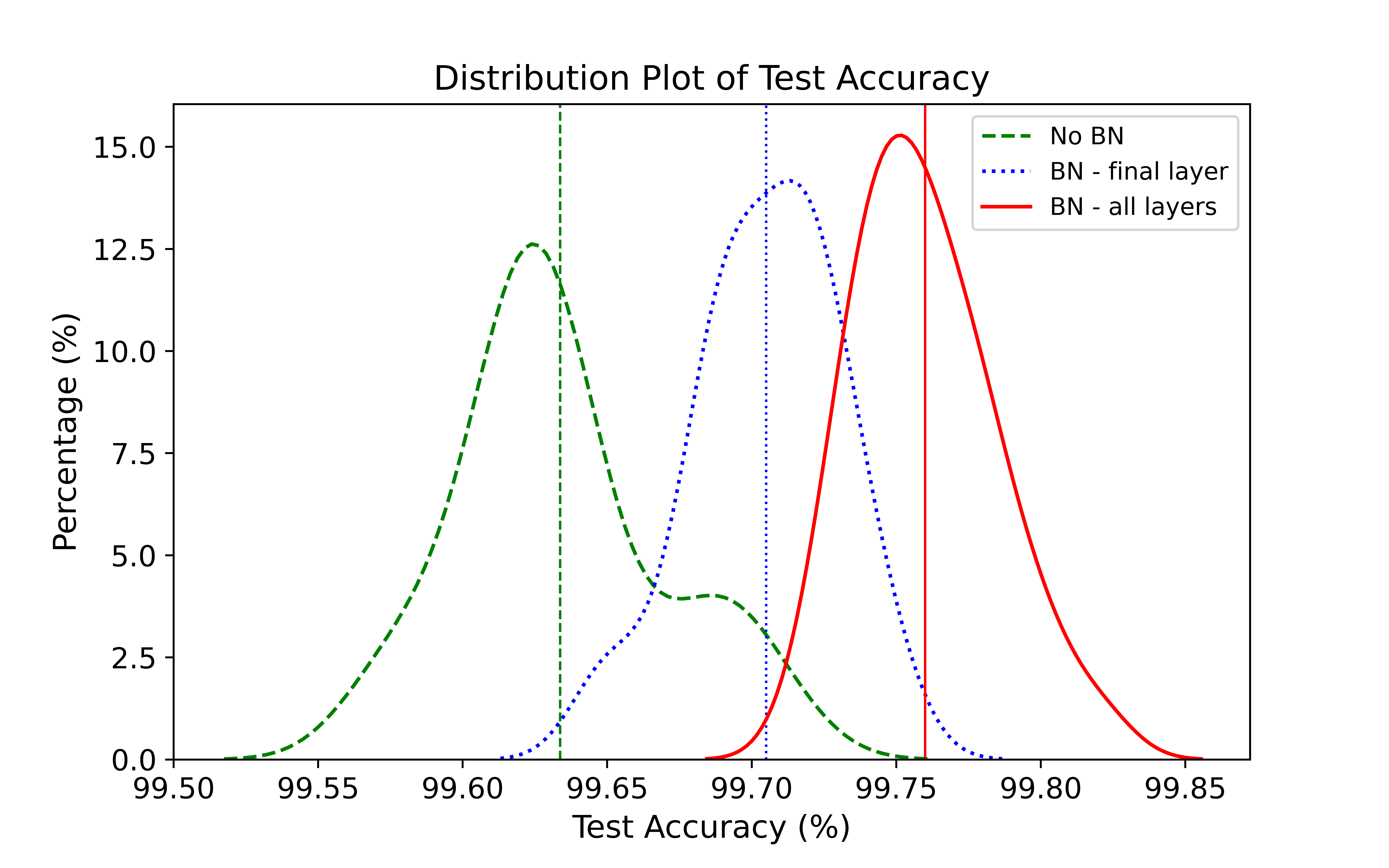}
	\caption{Distribution of test accuracy for networks trained with different batch normalization schemes.}
	\label{graph:bn}
\end{figure}

Batch normalization is a well known technique to improve performance of the network as well as stability and speed of training \cite{ioffe15arxiv}. It has been reported that most neural network models benefit from using batch normalization \cite{bjorck18nips,santurkar18nips}. In this section we study the impact of batch normalization on the performance of the network model $M_5$. We compared three configurations: the first model uses no batch normalization at all, the second model uses batch normalization only at the fully connected layer, and the third model uses batch normalization at all layers. Figure \ref{graph:bn} shows the distribution plot of 30 networks for each configuration, and Table \ref{tab:bn} shows the 95\% confidence range of test accuracy for each configuration. It is evident that using batch normalization helps improve the performance of neural network models. The best performance is achieved when batch normalization is used at each convolution and fully connected layer.

\begin{table}[ht]
 \caption{95\% confidence range of test accuracy for networks trained with different batch normalization schemes.}
  \centering
  \begin{tabular}{C{2.0in}C{1.6in}}
    \toprule
    configuration              & test accuracy                 \\
    \toprule
    no batch normalization               & 99.6337 $\pm$ 0.0131  \\
    \midrule
    batch normalization at the final layer   & 99.7050 $\pm$ 0.0092  \\
    \midrule
    batch normalization at all layers    & \textbf{99.7600} $\pm$ \textbf{0.0089}  \\
    \bottomrule
  \end{tabular}
  \label{tab:bn}
\end{table}

\section{Conclusion}
The MNIST handwritten digit data set is often used as an entry-level data set for training and testing neural networks. While achieving 99\% accuracy on the test set is rather easy, correctly classifying the last 1\% of the images is challenging. People have tried many different network models and techniques to increase test accuracy, and the best accuracy reported reaches approximately 99.8\%. In this paper we showed that a simple CNN model with batch normalization and data augmentation could reach the best accuracy. Using an ensemble of homogeneous and heterogeneous network models could boost the performance, up to 99.91\% test accuracy which is one of the state-of-the-art performance. Studies with various different configurations show that the high performance is not achieved by a single technique or model architecture, but is contributed by multiple techniques such as batch normalization, data augmentation, and ensemble methods. 

\bibliographystyle{unsrt}  


\end{document}